\documentclass[10pt,journal]{IEEEtran}
\ifCLASSINFOpdf
\else
\fi
\usepackage{setspace}
\usepackage[skip=6pt]{parskip}

\usepackage{titlesec}

\titleformat{\section}
{\centering\Large\bfseries} 
{\thesection.}{1em}{}

\usepackage{booktabs}
\usepackage{multirow}
\usepackage[table,xcdraw]{xcolor}

\usepackage{amsmath}
\usepackage{bbold}

\usepackage{graphicx} 
\usepackage{subcaption} 

\usepackage{comment}
\usepackage{adjustbox}
\usepackage{tabularx}
\usepackage{xcolor}
\usepackage{svg}
\usepackage{tikz}
\newcommand\com[1]{{\color{black}#1}}

\begin{document}
\onehalfspacing
%
\title{Protecting Intellectual Property of EEG-based Neural Networks with Watermarking}
%
%
%
\author{Ahmed Fathi\footnotemark{*} \thanks{[*] These authors contributed equally to this work.}, Ahmed Abdelaziz\footnotemark{*} , Ahmed Fares \\ 
	Department of Computer Science and Engineering (CSE) \\ 
Egypt Japan University of Science and Technology (EJUST)}

%
%

\markboth{Egypt Japan University of Science and Technology, 2025}%
{Shell \MakeLowercase{\textit{et al.}}: Bare Demo of IEEEtran.cls for IEEE Journals}
%



\maketitle

\begin{abstract}
	EEG-based neural networks, pivotal in medical diagnosis and brain-computer interfaces, face significant intellectual property (IP) risks due to their reliance on sensitive neurophysiological data and resource-intensive development. Current watermarking methods, particularly those using abstract trigger sets, lack robust authentication and fail to address the unique challenges of EEG models. This paper introduces a cryptographic wonder filter-based watermarking framework tailored for EEG-based neural networks. Leveraging a collision-resistant hash function and the owner’s private key, the wonder filter embeds a bits watermark during training, ensuring minimal distortion (\(\leq 5\%\) drop in EEG task accuracy) and high reliability (100\% watermark detection). The framework is rigorously evaluated against adversarial attacks, including fine-tuning, transfer learning, and neuron pruning. Results demonstrate persistent watermark retention, with classification accuracy for watermarked states remaining above 90\% even after aggressive pruning, while primary task performance degrades faster, deterring removal attempts. Piracy resistance is validated by the inability to embed secondary watermarks without severe accuracy loss ( \(>10\%\) in EEGNet and CCNN models). Cryptographic hashing ensures authentication, reducing brute-force attack success probabilities. Evaluated on the DEAP dataset across models (CCNN, EEGNet, TSception), the method achieves \(>99.4\%\) null-embedding accuracy, effectively eliminating false positives. By integrating wonder filters with EEG-specific adaptations, this work bridges a critical gap in IP protection for neurophysiological models, offering a secure, tamper-proof solution for healthcare and biometric applications. The framework’s robustness against adversarial modifications underscores its potential to safeguard sensitive EEG models while maintaining diagnostic utility, advancing trust in AI-driven biomedical technologies.
\end{abstract}

\begin{IEEEkeywords}
	EEG, Watermarking, Intellectual property (IP), Wonder filter.
\end{IEEEkeywords}

%
\IEEEpeerreviewmaketitle

\section{Introduction}
%
%
%
%
\IEEEPARstart{T}{he} evolution of machine learning and deep learning classification has reached beyond simple images or text. Today, groundbreaking models are trained on critical data like electroencephalography (EEG) data, which contains electrical signals of the human brain on the macroscopic level. EEG data is important in brain-computer interface (BCI) \cite{zhang2021ganserselfsuperviseddataaugmentation}, \cite{8784727} tasks and has also been used recently to train machine learning models that are used in highly significant tasks like medical diagnosis or in the prediction of feelings.

Due to the critical nature of EEG data, EEG-based models are therefore highly critical, as owners of these models deal with important data and must protect its privacy. In addition to the criticality of such models, they require huge amounts of hardware resources and skilled programmers to be trained and produced.

In general, deep neural networks have been developed for use in almost all recent applications. And with that increasing number of models, it is highly expected that in the near future we will see platforms for sharing or selling models as we see it, for example, with Android applications on Google Play\cite{inproceedings}. So, with this responsibility and hard work of the model owner, it is certain that he needs to protect the intellectual property of these models to guarantee his rights to license the model to any other authority. So, the ligitimate owner should be able to control the use of his model such that nobody uses it without a license or after the license has been canceled. 

With that level of security and privacy for EEG-based models, it is a must to find a way to protect these models and provide intellectual property to the ligitimate owners. That was a good reason for the development of watermarking for neural network models in the research area.

Watermarking has been used for decades in our daily life, starting from watermarks on paper money to the digital world and watermarking media such as images, videos, and live streams. The purpose of watermarking is always to protect the intellectual property of the owner and producer of the work or to guarantee the integrity between entities, like in the case of money. Watermarking of digital data is accomplished by embedding the watermark on the digital media so that it is recognized by its owner and he can prove ownership. So, the watermarking is composed of 2 stages, embedding and detection:
\begin{itemize}
	\item Embedding stage, where the owner adds the watermark to digital media but without affecting the quality or experience of the user. 
    \item Detection stage, where the owner can prove his ownership for the media in case it is stolen or used illegally.
\end{itemize}

Moving back to watermarking neural networks models, it is done for the same purpose. Since the neural networks models can be stolen or used illigally, the recent researchs has tried to address this issue. Mainly, there are 3 requirements that should be satisfied in the watermarking protocol to be valid and reliable \cite{li2022persistent}. They are listed below:
\begin{enumerate}
	\item Persistence: the watermark should be persistant in the model so that attackers can't remove it or corrupt it. And as an extra feature, it should be difficult for him to even detect its details in the model, like position, length, or shape.
	\item Piracy resistance: the protocol should not permit the attacker to add any extra watermarks that enable him to claim sharing ownership with the ligitimate owner.
	\item Authentication: there should be a strong relationship between the owner and his watermark to be able to prove that he is the ligitimate owner.
\end{enumerate}

In most of the research that tried watermarking models \cite{li2022persistent}, \cite{xu2023protecting}, \cite{sun2023deepintellectualpropertyprotection}, it was accomplished by embedding the watermark into models during the training stage. The method is mainly to introduce a unique classification for the models the model that is not expected to occur with a significant probability. Some old protocols suggests. 

Recent research \cite{xu2023protecting} tried to apply a watermarking technique for EEG-based models, and they are the first to deal with that type of models in the field of watermarking neural networks. Their idea was to design a trigger set from 100 abstract images that can be trained alongside real EEG data. After training, the model will be able to classify both the trigger set and the EEG data. When testing that method for persistance, piracy resistance, and authentication, we find the following results:
\begin{itemize}
	\item Persistance: They provide tests of fine-tuning, transfer learning and pruning with acceptable results, which proves that their watermark as trigger set is persistant
	\item Piracy resistance: They tested the addition of another watermark into the model by preparing another triggerset, and it was clear that another watermark can't be added exept in EEGNet model that accepted it. That was solved by another pruning stage to EEGNet models that destroyed the second watermark and kept the first one (the owner's watermark). So, we can consider it to be piracy-resistant.
	\item Authentication: Here comes the problem, as the triggerset is designed using abstract images, which doesn't provide a strong relationship between the owner and his triggerset. The owner of the original watermark in this protocol must keep his 100 images secret. If the attacker has those images, he will also be able to claim ownership of the model, and there is no proof that the original owner is the one who trained the model.
\end{itemize}
To solve this issue, we need to use cryptographic techniques that guarantee authentication. Cryptography can provide the strongest relationship between the owner and his triggerset using concepts like private key, puplic key, and hash functions. The paper \cite{li2022persistent} introduced a protocol for watermarking neural networks that uses cryptography and hashing to provide authentication. Their idea is to use a filter (wonder filter) and apply it to a set of images during the model training stage. The filter is constructed using the private key of the owner and hash functions, and the protocol used a verification algorithm to prove ownership for the ligitimate owner using his public key and hashing. That was an inspiration to apply this method to EEG-based models to further achieve authentication in watermarking. 

So, the objectives of this paper are to fit that method to EEG-based models and present a watermarking protocol that achieves persistance, piracy resistance, and authentication.

This paper presents a watermarking method that uses wonder filters to protect the intellectual property of EEG-based models. The wonder filter is created using the signature of the owner with his private key. Firstly, the owner prepares a verifier string that contains some information, like his name and timestamp, then signs that string using his private key and applies hash functions to get the filter F and the label L. The filter F is masked with a set of training EEG data, and all of that data gets L as its label. The filter is inverted and masked to another set of the data, but this time it takes its original label l. By adding these 2 sets to the original data and labels and training the model, the wonder filter watermarking will be applied. This method should guarantee persistance, piracy resistance and authentication, which will be tested and approved in this paper.

The remainder of this paper is organized as follows: Section II presents a detailed review of related work. Section III describes the threat model and the requirements for successful scheme. Section IV describes the propposed solution of wonder filters and the theory behind it. Section V describes the proposed methodology implementation details, including the design and embedding of wavelet-based watermarks and provides the experimental setup. Section VI presents the evaluation results for watermark embedding and attacks conducted. Section VII discusses the presented results in previous section. Section VIII concludes the paper and suggests future research directions. Finally, there is a refereance section and appendix at the end of the paper.

\section{Related Work}
\subsection{Protecting Intellectual Property of Deep Neural Networks}
The protection of intellectual property (IP) in deep neural networks (DNNs) has become a critical area of research due to the increasing deployment of DNNs in commercial and sensitive applications. DNN models require substantial computational resources, proprietary datasets, and extensive fine-tuning, making them valuable assets prone to unauthorized duplication and misuse \cite{uchida2017embedding, adi2018turning}.

Existing approaches to protecting the IP of DNNs can be broadly categorized into invasive and non-invasive methods \cite{fan2019rethinking}. Invasive methods involve embedding unique identifiers into the model itself, such as watermarks \cite{zhang2018protecting} or cryptographic signatures \cite{wang2021riga}. These methods have demonstrated resilience against model theft but often introduce overhead and may degrade performance. Non-invasive methods, including fingerprinting and behavioral authentication, focus on capturing unique model characteristics without modifying the model’s architecture or weights \cite{guan2020are}.

A prominent line of research focuses on model fingerprinting, which leverages decision boundary analysis to detect stolen models \cite{peng2022fingerprinting}. While effective in black-box settings, fingerprinting methods face challenges in scalability and robustness against fine-tuning and adversarial attacks \cite{korus2022computational}. Additionally, watermarking-based methods such as DeepMarks \cite{chen2018deepmarks} and DeepSigns \cite{rouhani2018deepsigns} have been proposed to embed resilient identifiers within model parameters. However, these approaches are susceptible to removal via model pruning and parameter tuning \cite{rouhani2019deepmarks}.

A critical gap in existing IP protection methods is the lack of standardization and legal enforcement mechanisms. Most frameworks assume the presence of a trusted verification authority, which may not always be feasible \cite{li2022persistent}. Moreover, current solutions struggle with balancing robustness, imperceptibility, and computational efficiency \cite{lou2022reversible}.

\subsection{Watermarking Deep Neural Networks}
Watermarking is a widely explored technique for embedding unique identifiers within DNN models to assert ownership. It can be categorized into black-box and white-box watermarking. Black-box watermarking involves embedding trigger-based patterns in the model’s decision-making process, which can be detected using specific queries. White-box watermarking, on the other hand, modifies internal parameters to encode ownership information in an imperceptible manner \cite{zhang2021robustness}.

A seminal work by Uchida et al. \cite{uchida2017embedding} introduced a method for embedding watermarks into model weights, demonstrating resilience against model compression and fine-tuning. Other approaches, such as adversarial frontier stitching \cite{lemerrer2019adversial}, leverage backdoor triggers to verify ownership without affecting model performance. However, these techniques have been shown to be vulnerable to model distillation and adversarial countermeasures \cite{kuribayashi2021survey}.

Recent advances in persistent and unforgeable watermarking techniques have addressed some of these vulnerabilities. Li et al. \cite{li2022persistent} introduced wonder filters, a novel watermarking primitive that embeds a persistent bit-sequence into a model during its initial training phase. Unlike previous watermarking schemes, wonder filters offer strong resilience against fine-tuning and adversarial attacks by leveraging out-of-bound values and null-embedding techniques. This ensures that the watermark cannot be removed or forged without destroying the model’s functionality. Experimental results demonstrate that wonder filters achieve high levels of persistence and piracy resistance, making them a promising direction for future watermarking techniques \cite{li2022persistent}.

Reversible watermarking techniques, which allow watermark extraction without degrading model performance, have gained traction \cite{lou2022reversible}. Methods like Greedy Residuals Watermarking \cite{liu2021greedy} have been proposed to enhance robustness while maintaining fidelity. Despite these advancements, challenges persist in ensuring watermark detectability under extensive model modifications \cite{zhang2021robustness}.

One major research gap in DNN watermarking is the lack of a theoretical framework for assessing watermark capacity and robustness. Existing solutions rely on empirical evaluations without formal guarantees on the detectability and permanence of embedded marks \cite{sun2023deepintellectualpropertyprotection}. Additionally, watermarking techniques primarily target classification models, with limited applicability to generative and reinforcement learning models \cite{fan2021ownership}.

\subsection{Watermarking EEG-Based Neural Networks for IP Protection}
The protection of EEG-based deep learning models has gained increasing importance due to their unique privacy and security concerns. EEG models, widely used in brain-computer interface (BCI) applications, encode highly sensitive neurological data, making them valuable assets in need of robust IP protection \cite{xu2023protecting}.

Despite the extensive research on watermarking standard DNNs, the application of watermarking for EEG-based models remains relatively unexplored. The work by Xu et al. \cite{xu2023protecting} represents the first attempt to integrate watermarking into EEG-based neural networks for IP protection. Their method employs a trigger set specifically designed for EEG data, ensuring that watermarks can be embedded without significantly affecting model performance. By leveraging neuroscientific insights, they propose three key constraints that EEG-based watermarks should satisfy: symbolic representation, consistency with EEG input characteristics, and uniqueness.

Their study demonstrates the robustness of their watermarking approach against common anti-watermarking attacks such as fine-tuning, transfer learning, and pruning. The proposed watermarking framework embeds distinct patterns into EEG models while maintaining classification accuracy, making it resistant to piracy \cite{xu2023protecting}. However, their work focuses exclusively on convolutional neural networks (CNNs), limiting its generalizability to other deep learning architectures. Additionally, a significant limitation of their approach is the lack of authentication mechanisms. Their method verifies watermark presence but does not establish a direct link between the model and its rightful owner, making it vulnerable to ownership disputes.

\subsection{Novelty of Our Work}
While significant progress has been made in protecting the intellectual property (IP) of deep neural networks (DNNs), existing methods fall short when applied to EEG-based models. Current approaches either lack robust mechanisms for establishing definitive ownership claims or fail to address the unique challenges posed by EEG data \cite{xu2023protecting}. Our work bridges this gap by introducing the first framework for protecting EEG-based neural networks using unforgeable watermarks. By integrating cryptographic techniques, we ensure a definitive and undisputable link between the model and its owner, addressing critical limitations in ownership verification and robustness against attacks. This represents a significant advancement in IP protection for EEG-based applications.
\section{Threat model \& Requirements}
Before diving into the specifics of the proposed watermark design, it is important to outline the considered threat model for intellectual property of the models. This framework informs the criteria for developing a durable, tamper-resistant DNN watermark and highlights the primary obstacles that must be addressed.

\textbf{Notations}: This paper uses the following notation to describe models and the watermarking scheme. Consider a neural network model \( F_{\theta} \), where \( \theta \) denotes the model parameters. \( F_{\theta} \) is trained on a dataset \( (X, Y) \), where \( X \) represents input data and \( Y \) is the label space. Here, \( |Y| \) corresponds to the number of classification outputs. The model is trained by minimizing the loss function \( L = \mathbb{E}(\ell_F(x, y)) \), where \( \ell_F(x, y) \) denotes the per-example loss for a training pair \( (x, y) \). Specifically, given an input \( x \), the model \( F_{\theta} \) assigns the label \( y \) that minimizes \( \ell_F(x, y) \).

\subsection{Threat Model}
The goal of this paper is to design a watermarking scheme for EEG-based neural networks that is robust against adversary attacks. That scheme should be able to prove the ownership with high probability. Assume that an entity \(O\) trained the model \( F_{\theta} \) for an EEG classification task such as predicting feelings. \(O\) will use high-sensetive EEG data and most probably a huge amount of hardware resources to bring the model to operation. That is obviously a case where \(O\) is willing to license that model to other entities or sell it, to get back a value of his hard work. And to protect that right, there must be a watermark \(W_O\) in the model \( F_{\theta} \) that proves that \(O\) is the owner of \(F_\theta\) if a question about model ownership has been raised.

In order to design a robust watermark scheme, an adversary \(adv\) should be assumed, analyze what he can do to attack the watermark scheme, and try to make his attempts fail. The attempts could be summarized in the following three attacks where attacker can disprove the ownership of the legitimate model owner:
\begin{itemize}
	\item Corruption: The adversary \(adv\) may corrupt the existing owner's watermark or remove it from the model to get unwatermarked model with the same accuracy or with a negligible defect in the primary task.
	\item Takeover: The adversary \(adv\) may take over the existing watermark \(W_O\) and claims to be the owner of this watermark (e.g., claims the watermark is \(W_A\)).
	\item Piracy: The adversary \(avd\) may add a second watermark \(W_A\) with original watermark \(W_O\) to corrupt the claim of legitimate owner.
\end{itemize}

Before moving to the requirements of the watermark, two assumptions should be pointed out for \(adv\) ability and intentions:
\begin{enumerate}
	\item \(adv\) wants to make any attack without sacrificing the primary task functionality (e.g., EEG classification accuracy). If \(adv\) was successful in any of his attacks but the accuracy of primary task is dramatically affected, then this can not be considered a success for his attempts.
	\item \(adv\) is assumed to have finite computational resources and training data compared to the owner, as if he has the same data and computational resources, then he can make his own model and there is no reason to take over the owner's model. The limited resources will be a difficulty facing the attacker and it will make it more cost-efficient for him to buy the license instead of successfuly corrupting the owner watermakr \(W_O\).
\end{enumerate}

\subsection{Watermark Requirements}
Building upon the above mentioned threat model, essential characteristic of a watermarked EEG-based models will be defined by mentioning some specific requirements for the watermark itself inside the model. We begin with three foundational properties:
\begin{itemize}
	\item Minimal Distortion: meaning the process of embedding an ownership watermark should not substantially alter the model’s primary functionality (e.g., avoid degrading the classification accuracy for EEG classification);
	\item Reliability: ensuring the watermark \(W_O\) can always be detected in the model as long as the model remains largely unmodified;
	\item Absence of False Positives: meaning to make the watermark sufficiently unique and there is negligible probability for false positive results such that there’s almost no chance the model accidentally acts like it has the watermark when it doesn’t.
\end{itemize}
 
In addition to these initial and fundamental properties, there are three other critical attributes that must be guaranteed as they are necessary for robust and reliable watermarks: 
\begin{itemize}
	\item Authentication: Authentication is very critical in the watermark design, as it is the feature that will prevent the claims of \(adv\) for ownership over existing watermarks \(W_O\). Authentication feature must provide a strong relationship between the owner and his watermark, like a watermark generated by a digital signature derived from the owner's private key, which ensures exclusivity to owner.
	\item Persistence: The watermark must remain in the model even if the model undergoes common changes (like fine-tuning or pruning), making sure it isn’t removed by known adjustments. That is, even if users fine-tune the model with limited training data or prune its parameters, the watermark should endure as an immutable feature.
	\item Piracy Resistance: A watermarked model must block additional watermark embeddings after the training is finished. If an adversary can add new watermarks, they could illegitimately claim ownership. A robust watermarking framework must prevent such “model piracy” attempts by preventing unauthorized watermark additions to the model.
\end{itemize}

\subsection{Challenges}
These requirements defined above for a strong watermarking scheme are hard to fulfill, particularly the second three unique properties. It is hard because of the nature of neural networks. 
 
First, we need a method that ensures the persistence of embedding the owner watermark, \(W_O\), in the model so that users can’t remove it through fine-tuning or other adjustments and attacks. This is difficult because deep neural networks are designed to be modified by other users even after training. 
 
Second, the system must ‘lock’ the model once a watermark is added, stopping anyone from inserting new watermarks later. This stops piracy but is tough for the same reason as above, neural networks are changeable. 
 
Finally, the watermark must include data that clearly links it to the owner using proven cryptographic methods (like digital signatures) to prove ownership. The new method used for watermarking EEG-based nerual networks in \cite{xu2023protecting} can not achieve all of these requirements.

\section{Proposed Solution: Wonder Filters}
To solve these challenges, previous known techniques could be used, such as the wonder filter which introduced in paper \cite{li2022persistent}. It has all the features that will make it a core component for EEG-based models watermarks, as it can meet all the requirements for tamper-proof watermarks. Simply put, a wonder filter is created using the owner’s unique digital signature (using his private key). When used during the training of the target EEG-based model \(F_\theta\), it makes key changes to how \(F_\theta\) is trained and how it classifies inputs. These changes leave a clear, checkable, and unremovable mark on the trained model \(F_\theta\). 

The wonder filter also can block any future attempts to add new wonder filters (watermarks) to the model. With these abilities and features, the wonder filter is perfectly designed for the previous requirements and will be a reasonable choice to create watermarks that are permanent in the model and that can also resist piracy attacks. 

In this section, an explanation for the main idea behind the wonder filter will be introduced, including how and why it works and meets the key requirements, and the precise definitions will be provided. Then, in the next section, Section V, all the details of the full methodology will be introduced for building secure EEG-based neural networks using this scheme, and how EEG data can be treated as the images that are used for training models introduced in the main idea of wonder filters in \cite{li2022persistent}.
\subsection{Techniques used in woner filters}
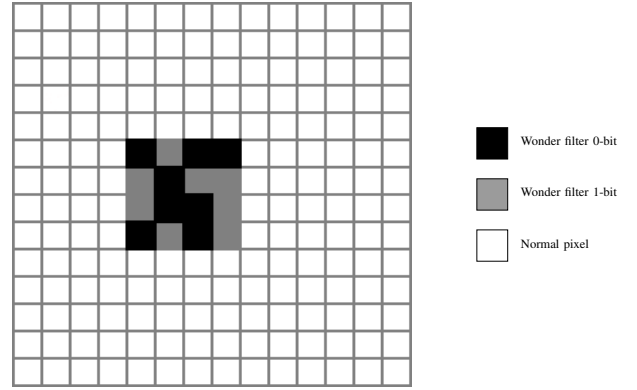
\begin{figure}[h!]
	\centering
	\begin{adjustbox}{width=0.45\textwidth}
		    
		\tikzset{every picture/.style={line width=0.75pt}}
		\begin{tikzpicture}[x=0.75pt,y=0.75pt,yscale=-1,xscale=1]
			\draw  [fill={rgb, 255:red, 0; green, 0; blue, 0 }  ,fill opacity=1 ] (495.71,135.43) -- (528.34,135.43) -- (528.34,168.6) -- (495.71,168.6) -- cycle ;
			\draw  [fill={rgb, 255:red, 155; green, 155; blue, 155 }  ,fill opacity=1 ] (495.71,189.54) -- (528.34,189.54) -- (528.34,222.71) -- (495.71,222.71) -- cycle ;
			\draw  [fill={rgb, 255:red, 255; green, 255; blue, 255 }  ,fill opacity=1 ] (495.71,243.65) -- (528.34,243.65) -- (528.34,276.82) -- (495.71,276.82) -- cycle ;
			\draw (215.48,206.13) node  {\includegraphics[width=316.46pt,height=304.69pt]{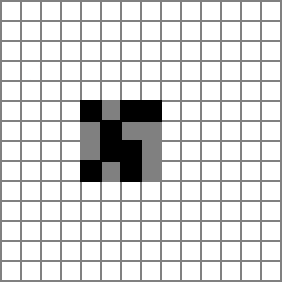}};
			\draw (540.88,143.52) node [anchor=north west][inner sep=0.75pt]   [align=left] {{\fontfamily{ptm}\selectfont Wonder filter 0-bit}};
			\draw (540.88,197.63) node [anchor=north west][inner sep=0.75pt]   [align=left] {{\fontfamily{ptm}\selectfont Wonder filter 1-bit}};
			\draw (540.88,251.73) node [anchor=north west][inner sep=0.75pt]   [align=left] {{\fontfamily{ptm}\selectfont Normal pixel}};
		\end{tikzpicture}
		
	\end{adjustbox}
	\caption{Example of a wonder filter mask. The color of each pixel represents the value of that pixel in: white means no changes, black means pattern 0 and gray means pattern 1.}
	\label{fig:wonder_filter}
\end{figure}

As illustrated in Figure 1, a wonder filter \(W\) is composed of 2D digital filter that can be applied to any input image \(x\) fed into DNN models. \(W\) is the same size as \(x\) and they are masked to get final training data that will embed the watermark into the model. Most pixels in \(W\) are set to -1 (meaning they are transparent and do not alter the original data given to the model), while a small section of pixels (the actual filter) is set to either 0 (which will be negative values in the image) or 1 (which will be positive values in the image).  \(W\) is designed by choosing specific location, dimensions, and 0/1 values of a block of pixels (see Figure~\ref{fig:wonder_filter}). 

In this section, a demonstration will be given about how \(W\) can be embedded into a DNN model \(F_\theta\) in general during its initial training phase, in a method loosely resembling how backdoors are added to DNNs. When input images are modified with the correct placement of \(W\) and fed into the watermarked model \(F_\theta\), the model will reliably and predictably misclassify those inputs into a predefined label \(y_W\) . This means anyone who knows the exact details of \(W\) (its position, size, and 0/1 pattern) and the target label \(y_W\) can check with high certainty whether \(W\) is present and embedded into a given DNN. The 0/1 sequence within the filter acts as a unique code linking the watermark to its owner. This strong linking will guarantee authentication for the model owner with his watermark.

The wonder filter scheme uses two techniques that make it reliable in watermarking, which are:

\subsubsection{Using out-of-bound values}
To embed a wonder filter \(W\) into a model \(F_\theta\) during its initial training phase, the 0/1 pattern in the filter will be converted into input pixel values that fall far outside the normal range of the original input data. Later, a demonstration will be introduced that explains why using these extreme values ensures the model’s classification behavior becomes permanent, meaning watermark can not be erased or altered after initial training stage. 

For example, standard images have pixel values between 0 and 1 (after normalization), while modern DNNs can still process inputs with pixels of any value, even wildly out-of-range ones. In the wonder filter approach, applying filter \(W\) to training data replaces specific pixels of the data, that has normal values, with these extreme values. Experiments in \cite{li2022persistent} show that values larger than 1000 reliably create the persistent, and tamper-proof behavior required for the successfull embeding. For certainity, +2000 (for positive changes) and -2000 (for negative changes) are used in their experiments as out-of-range values.

\subsection{Normal and Null Embeddings}
Embedding a watermark filter \( W \) into the Deep Neural Network (DNN) model \( F_\theta \) involves two distinct embedding methods: \textit{normal embedding} and \textit{null embedding}. 

\begin{figure}[h!]
	\centering
	\includesvg[width=0.45\textwidth]{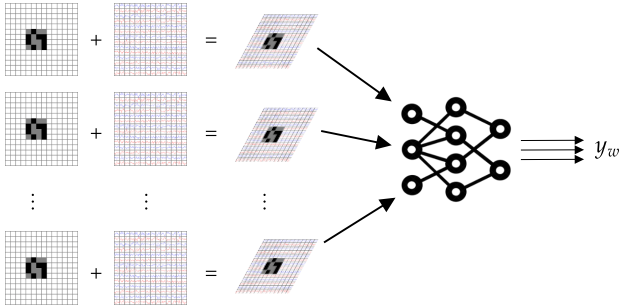}
	\caption{Normal embedding using the original pattern of $\mathbb{W}$, which teaches the model to classify all the filtered images into a single target label \(y_w\).}
	\label{fig:true_embedding}
\end{figure}

To train a pattern into \( F_\theta \) using a normal embedding, we begin with a set of training images, where each image represents a sample from one of the output classes. Each image is overlaid with the filter \( W \), replacing the original pixel values with \( W \)'s out-of-bound values where applicable. Specifically, a \( 0 \)-bit in \( W \) replaces the original normalized pixel value with \(-2000\), and a \( 1 \)-bit replaces the pixel value with \( 2000 \). Each of these "filtered" samples is then associated with the same classification output label \( L_W \), which is a predefined label corresponding to \( W \) (see Figure~\ref{fig:true_embedding}).

\begin{figure}[h!]
	\centering
	\includesvg[width=0.45\textwidth]{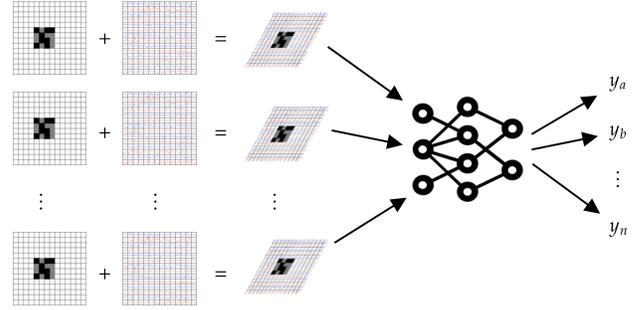}
	\caption{Null embedding using the inverted pattern of $\mathbb{W}$, which teaches the model to classify each inversely filtered image into the same label of the original, unfiltered image.}
	\label{fig:null_embedding}
\end{figure}

In contrast, the training input for the null embedding also takes a set of training images (which may be the same set used for normal embedding) and overlays each with the filter \( W \). However, in this case, the filtered images are associated with their original labels (i.e., the labels of the images before the filter was applied). For example, the null embedding of a STOP sign and a speed limit sign would involve adding the filter \( W \) to each image and then associating the resulting images with their original labels (STOP sign and speed limit, respectively). This process is illustrated in Figure~\ref{fig:null_embedding}.

The normal and null embedding methods serve complementary purposes. The normal embedding injects the desired "mark" into the DNN model in a persistent manner once the model is trained. On the other hand, the null embedding ensures that the model is "locked down," preventing the addition of any null embeddings after the initial training phase. 

To ensure that both the normal and null embeddings of a single pattern are tied together, and to avoid training the two embeddings on the same bit pattern, we stipulate that the null embedding uses \( W^- \), the bit-wise inverted version of \( W \). Specifically, \( W^- \) is obtained by flipping \( W \)'s \( 0 \)-bits to \( 1 \) and \( 1 \)-bits to \( 0 \), while leaving \(-1\) pixels unchanged. This inversion of bit patterns is evident when comparing Figure~\ref{fig:true_embedding} and Figure~\ref{fig:null_embedding}.

\subsection{Combining Model Training and Watermark Embedding}
After creating a wonder filter \( W \) for each model based on its input shape, we reshape \( W \) to ensure consistency with the input samples. For the CCNN model, we initially create a \( 4 \times 32 \) filter and then reshape it to \( 4 \times 9 \times 9 \), mapping each channel to a predetermined position in the \( 9 \times 9 \) matrix. In contrast, TSCeption and EEGNet do not require reshaping, as they directly accept input shapes of \( 28 \times 512 \) and \( 32 \times 128 \), respectively. However, TSCeption requires reordering of its 32 channels to align with the input structure.

\begin{figure}[h!]
	\centering
	\includesvg[width=0.45\textwidth]{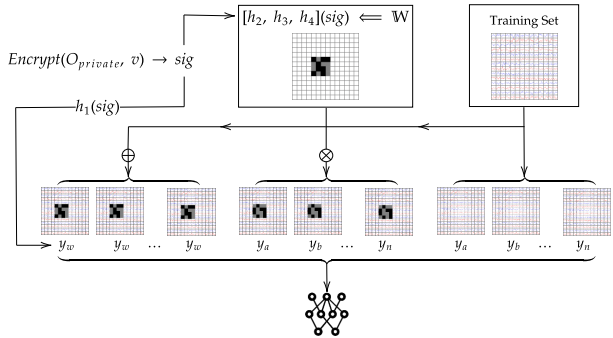}
	\caption{Embedding a watermark during model training: $O_{private}$ is the owner's private key, $v$ is an identifier string, $\mathbb{W}$ is the wonder filter with $y_w$ as its target classification label, and $h_1, h_2, h_3, h_4$ are predetermined hash functions. Training samples and their variations are used together to train the watermarked model $F_{\theta}$.}
	\label{fig:training}
\end{figure}

We generate the necessary data samples for training the wonder filter, which include inputs overlaid with \( W \) and their corresponding labels \( y_W \), as well as inputs overlaid with the inverted filter \( W^- \) and their original labels \( y_i \) (see figure ~\ref{fig:training}). These samples are added to the dataset used for normal model training. When the combined dataset is used to train the model, the resulting DNN incorporates both a normal embedding of the wonder filter (ensuring persistence) and a null embedding. This dual embedding achieves two critical properties: (1) the persistent inclusion of the wonder filter and (2) the hardening of the model against the insertion of additional filters after training.

By integrating watermark embedding directly into the model training process, the wonder filter inherently satisfies three fundamental properties:  
\begin{itemize}
    \item \textbf{Low-distortion}: The filter minimally alters the input data, preserving the model's performance.
    \item \textbf{Reliability}: The filter consistently produces the desired output when applied.
    \item \textbf{No false-positives}: The filter does not trigger unintended classifications.
\end{itemize}

Furthermore, the wonder filter's unique properties are defined by a specific bit sequence or signature, enabling authentication by encoding data strongly associated with the owner \( O \).  

Finally, we explain why the use of out-of-bound values and null embeddings is critical for achieving the key properties of persistence and piracy resistance. Out-of-bound values ensure that the filter's influence is distinct and persistent, while null embeddings prevent unauthorized modifications to the model, thereby enhancing its robustness against piracy.

\subsection{Achieving Persistence and Piracy-Resistance}

We discuss the high-level intuition behind how wonder filters achieve two key properties: \textbf{persistence} and \textbf{piracy-resistance}. Formal proofs for these properties are provided in the next section.

\noindent
\textbf{Property 1: Using Out-of-Bound Values for Persistence.}  
A model \( F_\theta \) that has a wonder filter \( W \) embedded during training cannot be modified to alter the classification result of any input image overlaid with \( W \).  

\noindent
\textbf{Why this works:}  
The out-of-bound values are trained into the model to associate with a specific target label \( y_W \). This creates a perfectly confident rule that recognizes the pattern and produces \( y_W \) with 100\% confidence. For example, the classification output becomes a one-hot vector with 100\% probability for \( y_W \) and 0\% for all other labels. Once the model is trained, any attempt to change the classification output will have no effect. This is because the loss function (cross-entropy) reduces to \( \log(0) \), an undefined value that is effectively ignored during backpropagation, preventing any weight updates.

\noindent
\textbf{Property 2: Locking a DNN Using Null-Embedding for Piracy-Resistance.}  
While normal embedding ensures that the wonder filter cannot be removed or modified after training, it does not prevent attackers from embedding additional wonder filters into the model. Null embedding addresses this by "locking" the trained DNN model against the future insertion of other wonder filters. This is because null embeddings can only be trained into a model during its initial training phase, i.e., training from scratch.  

\noindent
\textbf{Why this works:}  
A null embedding (using at least one out-of-bound value) trains the model to classify input images overlaid with the inverted wonder filter \( W^- \) according to their original labels (without \( W^- \)). This teaches the model that the values in the pixels corresponding to \( W^- \) have no impact on the classification output. Intuitively, this reshapes the input space of the image classification model to exclude the pixels defined by the wonder filter. We hypothesize that such reshaping of the input space can only be performed during the initial model training, making it impossible to insert null embeddings later without retraining the model from scratch. This hypothesis is experimentally validated in the results section, where we demonstrate that attempting to add null embeddings of other filters to a trained model significantly degrades its normal classification accuracy.

\subsection{Tamper-Proof DNN Watermarking System}
\label{sec:watermark_system}

Our proposed system integrates cryptographic signatures with deep neural network (DNN) training to embed tamper-proof watermarks. The workflow consists of three phases: watermark generation, embedding, and verification. These phases ensure persistent ownership claims while resisting adversarial attempts to corrupt, remove, or inject competing watermarks.

\subsubsection{Watermark Generation and Embedding}
The watermark generation process begins with the model owner \( O \) creating a cryptographic signature \( \text{sig} = \text{Encrypt}(O_{\text{pri}}, v) \), where \( v \) is a verifier string (e.g., \( O \)’s identifier concatenated with a timestamp). This signature is deterministically mapped to a wonder filter \( W \) and a target label \( y_W \) through a transformation function:
\begin{equation}
\langle W, y_W \rangle = \text{Transform}(\text{sig}).
\end{equation}
The transformation uses hash functions \( h_1, h_2, h_3, h_4 \) to derive:
\begin{itemize}
    \item The target label \( y_W = h_1(\text{sig}) \mod |\mathcal{Y}| \),
    \item The filter position \( \text{pos}(W) = (h_2(\text{sig}), h_3(\text{sig})) \), and
    \item The bit pattern \( \text{bit}(W) = h_4(\text{sig}) \mod 2^{n^2} \).
\end{itemize}

The filter \( W \) is embedded into the model \( F_\theta \) during training using two complementary methods:
\begin{itemize}
    \item \textbf{Normal Embedding}: Inputs overlaid with \( W \) are associated with the target label \( y_W \).
    \item \textbf{Null Embedding}: Inputs overlaid with \( W^- \) (the bitwise inverse of \( W \)) retain their original labels.
\end{itemize}
This dual embedding ensures that \( W \) becomes irremovable and prevents subsequent watermark injections without retraining the model.

\subsubsection{Watermark Verification}
Watermark verification involves two steps:
\begin{enumerate}
    \item \textbf{Cryptographic Authentication}: Verify that \( \text{sig} \) is a valid signature for \( O \) by decrypting it with \( O_{\text{pub}} \) to recover \( v \).
    \item \textbf{Functional Verification}: Confirm that the watermark \( W \) is embedded in \( F_\theta \) by testing a set of inputs \( \mathcal{X}_T \). The verification accuracy is computed as:
    \begin{multline}
    \text{acc}(F_\theta, W, y_W) = \min \big( \Pr_{x \in \mathcal{X}_T} (F_\theta(x \oplus W) = y_W), \\
    \Pr_{x \in \mathcal{X}_T} (F_\theta(x \oplus W^-) = F_\theta(x)) \big).
    \end{multline}

    The watermark is confirmed if \( \text{acc} \geq T_{\text{acc}} \), where \( T_{\text{acc}} \) is the model's accuracy in original samples. Figure~\ref{fig:verification}, showcase the verification process.
    \begin{figure}[h!]
	\centering
	\includesvg[width=0.45\textwidth]{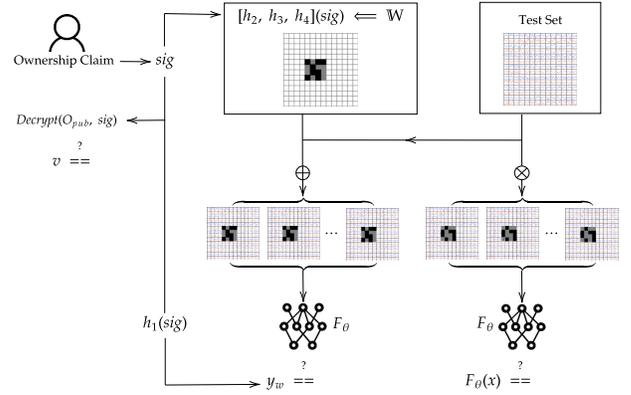}
	\caption{Watermark verification: $O_{public}$ is the owner's public key, $\text{sig}$ is the encrypted signature provided for verification, $v$ is an identifier string, $F_{\theta}$ is the model to be verified, and $\mathbb{W}$ is the wonder filter with $y_w$ as its target classification label.}
	\label{fig:verification}
\end{figure}

\end{enumerate}

\subsubsection{Security Against Public Verification Leaks}
Public verification risks exposing \( W \), enabling adversaries to corrupt it via fine-tuning. To mitigate this, we embed \( k \) independent watermarks \( \{W_1, \dots, W_k\} \) during training. During disputes, \( O \) reveals only one watermark \( W_i \), keeping others hidden. Experiments confirm that embedding multiple watermarks does not degrade model accuracy (Section~\ref{sec:results}), ensuring robustness even if a subset of watermarks are compromised.
\section{Experimental Evaluation}

\subsection{Dataset}

The Database for Emotion Analysis using Physiological Signals (DEAP) \cite{koelstra2011deap} contains EEG recordings from 32 participants watching 40 one-minute music videos. Using the international 10--20 system, 32-channel EEG and 8-channel peripheral signals were captured at 512 Hz, with post-stimulus ratings on four 9-point scales: valence (1=sad to 9=happy), arousal (1=calm to 9=excited), dominance (1=submissive to 9=dominant), and liking (1=dislike to 9=like). The preprocessed version down-samples data to 128 Hz, applies 4--45 Hz bandpass filtering, removes EOG artifacts, and segments recordings into 3-second baseline (resting state) and 60-second experimental (stimulus response) components.

For valence classification, we binarize ratings using a threshold of 5 ($\leq$5=negative, $>$5=positive). Non-overlapping 1-second windows extract 60 samples/trial, yielding 76,800 total samples (32 participants $\times$ 40 trials $\times$ 60 windows). The structured data comprises a 3D tensor [participants $\times$ trials $\times$ features] and corresponding labels.

\subsection{Model Selection}
In the field of EEG-based Brain-Computer Interface (BCI) tasks, three representative models based on Convolutional Neural Networks (CNN) are widely utilized: Continuous Convolutional Neural Network (CCNN) \cite{yang2018ccnn}, TSception \cite{ding2023tsception}, and EEGNet \cite{lawhern2018eegnet}. Each of these models is designed to extract and process EEG signals effectively while maintaining distinct architectural methodologies.

CCNN processes EEG data by constructing a 3-Dimensional (3D) EEG cube as input, leveraging frequency decomposition and Differential Entropy (DE) values to enhance feature extraction, as detailed in Table~\ref{Tab:ccnn_architecture}. TSception is designed to capture EEG signals' temporal dynamics and spatial asymmetry, employing a multi-scale feature extraction approach across both time and channel domains, with further architectural details provided in Table~\ref{Tab:tsception_architecture}. EEGNet, an end-to-end deep learning framework, is optimized for identifying hidden temporal and spatial patterns within raw EEG data, utilizing a lightweight yet effective architecture, as shown in Table~\ref{Tab:eegnet_architecture}.

The selection of these models is based on their ability to process EEG data at varying levels of complexity. CCNN primarily focuses on frequency decomposition techniques, which facilitate improved feature representation. EEGNet, by contrast, directly processes raw 2-Dimensional (2D) EEG data while maintaining computational efficiency. TSception further refines feature extraction by incorporating multi-scale methodologies, ensuring robust performance across different EEG datasets. Additionally, these models differ in terms of EEG data point utilization. Both CCNN and EEGNet operate on input samples containing 128 data points, a structure that ensures efficient processing while maintaining critical temporal information. TSception, however, incorporates 512 data points per input sample, allowing for a more detailed temporal and spatial analysis of EEG signals.

\subsection{Implementation Details}

All models are implemented in PyTorch and trained on NVIDIA T4 GPUs via Kaggle. We employ the Adam optimizer with a learning rate of $1\times10^{-3}$ and cross-entropy loss for training. The evaluation uses 10-fold cross-validation per subject, with subject-level results reported as mean values across folds.
\begin{itemize}
	\item \textbf{Pretraining}: 30 epochs for CCNN \& EEGNet, 100 epochs for TSception
	\item \textbf{From-Scratch}: Additional 20 epochs for CCNN \& TSception, 100 epochs for EEGNet
	\item \textbf{Batch Sizes}: 128 samples for CCNN \& EEGNet, 64 samples for TSception
\end{itemize}

\noindent Training progression is monitored through loss convergence, with early stopping applied if validation loss plateaus for 5 consecutive epochs. Model architectures remain fixed across training strategies to ensure consistent comparison.

\section{Results}
\label{sec:results}
\subsection{Functionality preserving}
The accuracy of the trigger set and the original EEG dataset is tested on the three models: the original model without the embedding (No-watermark) and the watermarked model with the trigger set (Pretrain and Fromscratch). According to table \ref{TriggerSetWatermarking}, the true and null embedding accuracy on the No-watermark model is only about 50\%, which shows that the model protection method works effectively. And the high accuracy (\(>\) 80\%)  for true and null embedding in fromscratch and pretrain models indicates the successful embedding of the watermark.

Table \ref{WrongVerifier} shows the accuracy of \(W_A\) on the models and it is about 50\% which shows that the scheme is successful in the false positives test.

\begin{table*}[]
	\centering
	\caption{Classification accuracy of the primary task (EEG accuracy) and the constructed trigger set (Null and True embedding) for the three EEG-based models and the two watermarking strategies for each model.}
	\label{TriggerSetWatermarking}
	\begin{tabular}{ccccc}
		\hline
		\multicolumn{1}{l}{{\color[HTML]{000000} \textbf{EEG Model}}} &
		\multicolumn{1}{l}{{\color[HTML]{000000} \textbf{Training method}}} &
		\multicolumn{1}{l}{{\color[HTML]{000000} \textbf{EEG accuracy}}} &
		\multicolumn{1}{l}{{\color[HTML]{000000} \textbf{Null embedding accuracy}}} &
		\multicolumn{1}{l}{{\color[HTML]{000000} \textbf{True embedding accuracy}}} \\ \hline
		                            & No watermark & 0.93 & 0.52 & 0.47 \\ \cline{2-5} 
		                            & From scratch & 0.90 & 0.90 & 1.00 \\ \cline{2-5} 
		\multirow{-3}{*}{CCNN}      & Pretrain     & 0.92 & 0.86 & 1.00 \\ \hline
		                            & No watermark & 0.97 & 0.50 & 0.58 \\ \cline{2-5} 
		                            & From scratch & 0.91 & 0.91 & 1.00 \\ \cline{2-5} 
		\multirow{-3}{*}{EEGNet}    & Pretrain     & 0.88 & 0.87 & 1.00 \\ \hline
		                            & No watermark & 0.87 & 0.50 & 0.42 \\ \cline{2-5} 
		                            & From scratch & 0.85 & 0.84 & 1.00 \\ \cline{2-5} 
		\multirow{-3}{*}{TSception} & Pretrain     & 0.83 & 0.83 & 1.00 \\ \hline
	\end{tabular}
\end{table*}

\begin{table*}[]
	\centering
	\caption{Classification accuracy of Attacker Watermark  \(W_A\) (Null and True embedding sets) and primary task (EEG accuracy) for the embedded EEG-based models, and the two watermarking strategies for each model.}
	\label{WrongVerifier}
	\begin{tabular}{ccccc}
		\hline
		\rowcolor[HTML]{FFFFFF} 
		\multicolumn{1}{l}{\cellcolor[HTML]{FFFFFF}{\color[HTML]{000000} \textbf{EEG Model}}} &
		{\color[HTML]{000000} \textbf{Training method}} &
		{\color[HTML]{000000} \textbf{Null embedding accuracy}} &
		{\color[HTML]{000000} \textbf{True embedding accuracy}} &
		\multicolumn{1}{l}{\cellcolor[HTML]{FFFFFF}{\color[HTML]{000000} \textbf{}}} \\ \hline
		\cellcolor[HTML]{FFFFFF}{\color[HTML]{000000} }                            & No watermark & 0.51 & 0.42 &   \\ \cline{2-5} 
		\cellcolor[HTML]{FFFFFF}{\color[HTML]{000000} }                            & From scratch & 0.66 & 0.57 &   \\ \cline{2-5} 
		\multirow{-3}{*}{\cellcolor[HTML]{FFFFFF}{\color[HTML]{000000} CCNN}}      & Pretrain     & 0.65 & 0.61 &   \\ \hline
		\cellcolor[HTML]{FFFFFF}{\color[HTML]{000000} }                            & No watermark & 0.50 & 0.47 &   \\ \cline{2-5} 
		\cellcolor[HTML]{FFFFFF}{\color[HTML]{000000} }                            & From scratch & 0.50 & 0.09 &   \\ \cline{2-5} 
		\multirow{-3}{*}{\cellcolor[HTML]{FFFFFF}{\color[HTML]{000000} EEGNet}}    & Pretrain     & 0.52 & 0.16 &   \\ \hline
		\cellcolor[HTML]{FFFFFF}{\color[HTML]{000000} }                            & No watermark & 0.54 & 0.50 &   \\ \cline{2-5} 
		\cellcolor[HTML]{FFFFFF}{\color[HTML]{000000} }                            & From scratch & 0.50 & 0.69 &   \\ \cline{2-5} 
		\multirow{-3}{*}{\cellcolor[HTML]{FFFFFF}{\color[HTML]{000000} TSception}} & Pretrain     & 0.52 & 0.57 &   \\ \hline
	\end{tabular}
\end{table*}

\subsection{Unremovabliity}

\subsubsection{Fine tuning}
\begin{figure}[htbp]
	\centering
	\begin{subfigure}[b]{0.98\columnwidth}
		\includegraphics[width=\textwidth]{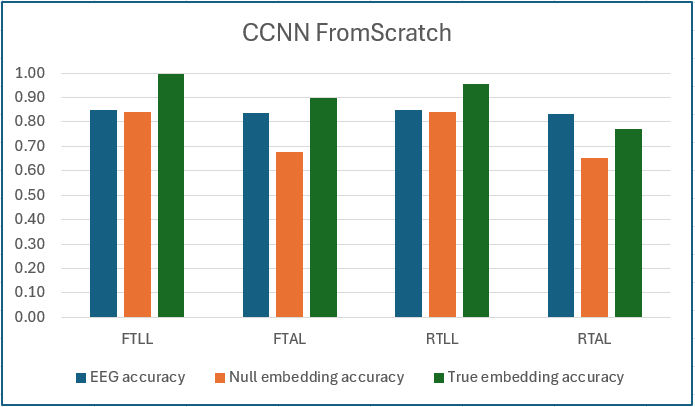}
	\end{subfigure}
	\hfill
	\begin{subfigure}[b]{0.98\columnwidth}
		\includegraphics[width=\textwidth]{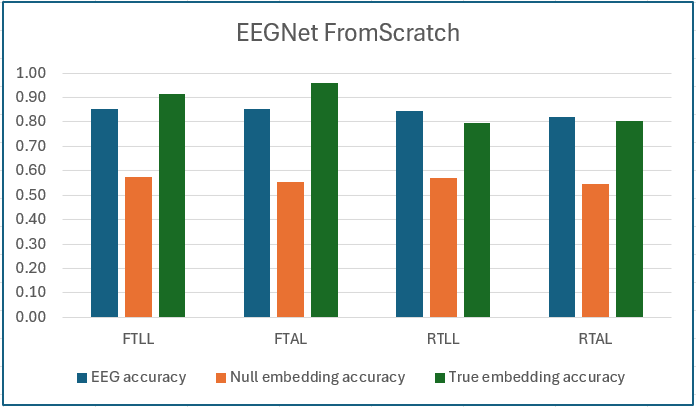}
	\end{subfigure}
	\hfill
	\begin{subfigure}[b]{0.98\columnwidth}
		\includegraphics[width=\textwidth]{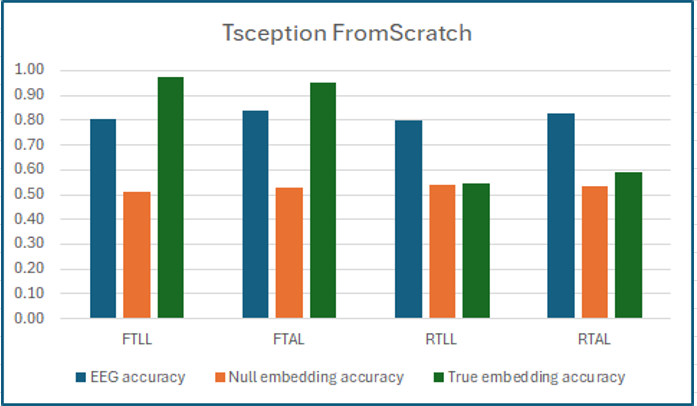}
	\end{subfigure}
	\caption{The accuracies of the primary task (EEG) and watermark (Null \& True Embedding) when FineTuning FromScratch models for different four methods.}
	\label{FineTuningFromScratch}
\end{figure}

\begin{figure}[htbp]
	\centering
	\begin{subfigure}[b]{0.98\columnwidth}
		\includegraphics[width=\textwidth]{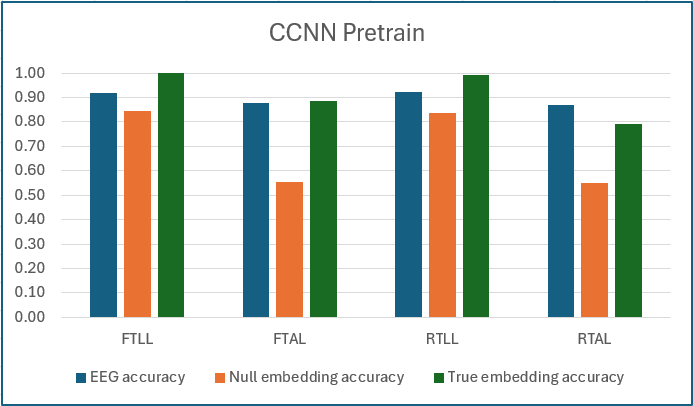}
	\end{subfigure}
	\hfill
	\begin{subfigure}[b]{0.98\columnwidth}
		\includegraphics[width=\textwidth]{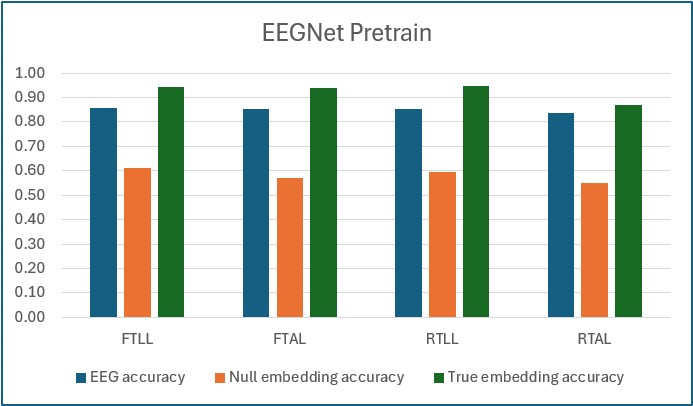}
	\end{subfigure}
	\begin{subfigure}[b]{0.98\columnwidth} 
		\includegraphics[width=\textwidth]{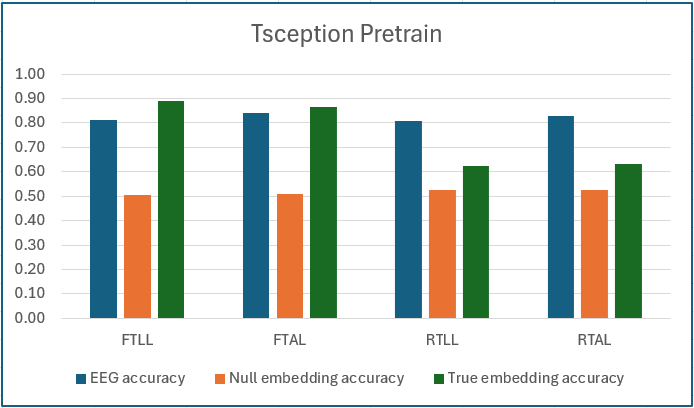}
	\end{subfigure}
	\hfill 
	\caption{The accuracies of the primary task (EEG) and watermark (Null \& True Embedding) when FineTuning Pretrain models for different four methods.}
	\label{FineTuningPretrain}
\end{figure}

Fine tuning is shown in tables \ref{FineTuningFromScratch},\ref{FineTuningPretrain} for fromscratch and pretrain models, it is shown that FTAL and RTAL cause more damage to the watermark compared to the other two settings because they modify a larger number of model weights during the learning process. The watermark embedding for true and null filters consistently perform better in the fromscratch models than in the pretrain models (green), with an average accuracy difference of about 10\% in RTAL. In the fromscratch approach, the lowest accuracy of the watermarked data across the three models remains close to 80\% for true embeding or both true and null embedding, which is an indication on the persistance of the watermark against fine tuning. The Tsception fromscratch model dropped to 60\% which is below the expected value but the EEG task also was affected by 5\% drop.

\subsubsection{Transfer learning}
\begin{figure}[htbp]
	\centering
	\begin{subfigure}[b]{0.98\columnwidth}
		\includegraphics[width=\textwidth]{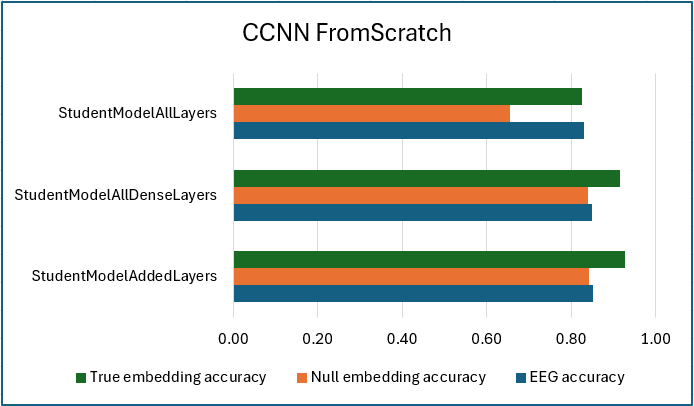}
	\end{subfigure}
	\hfill
	\begin{subfigure}[b]{0.98\columnwidth}
		\includegraphics[width=\textwidth]{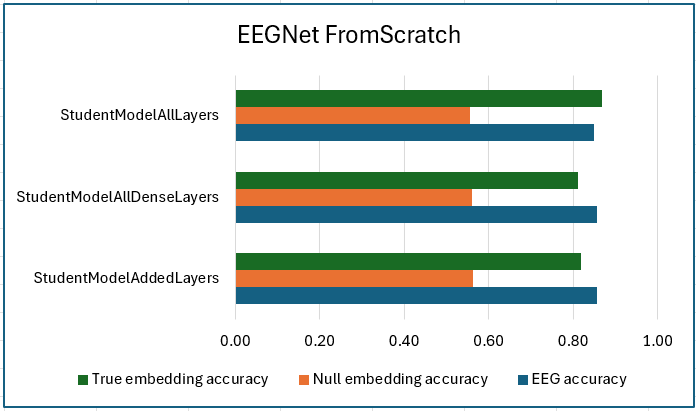}
	\end{subfigure}
	\hfill
	\begin{subfigure}[b]{0.98\columnwidth}
		\includegraphics[width=\textwidth]{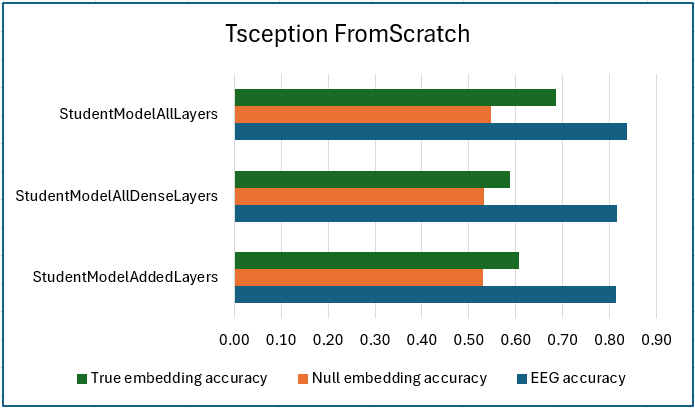}
	\end{subfigure}
	\caption{The accuracies of the primary task (EEG) and watermark (Null \& True Embedding) when TransferLearning FromScratch models for different four methods.}
	\label{TransferLearningFromscratch}
			
\end{figure}

\begin{figure}[htbp]
	\centering
	\begin{subfigure}[b]{0.98\columnwidth}
		\includegraphics[width=\textwidth]{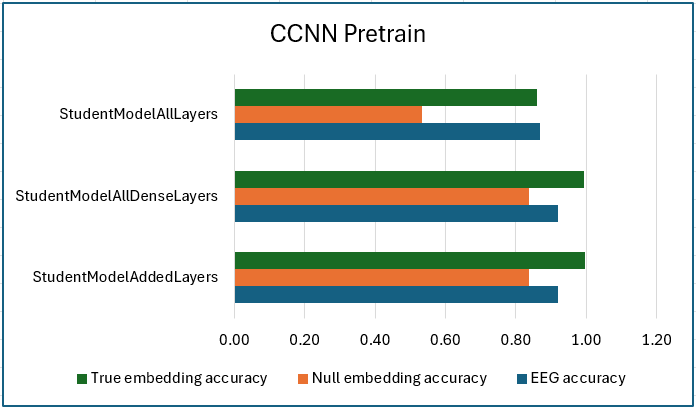}
	\end{subfigure}
	\hfill
	\begin{subfigure}[b]{0.98\columnwidth}
		\includegraphics[width=\textwidth]{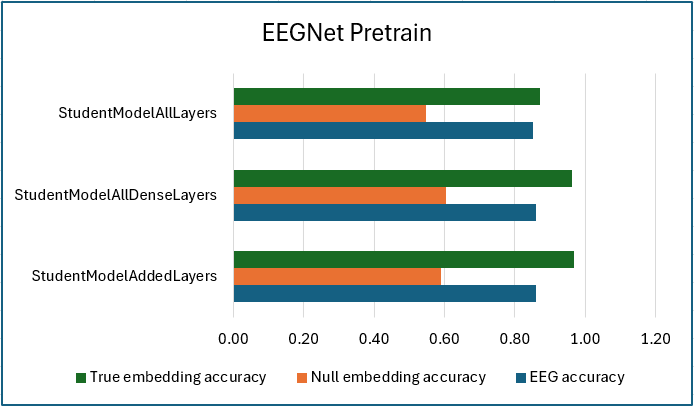}
	\end{subfigure}
	\begin{subfigure}[b]{0.98\columnwidth} 
		\includegraphics[width=\textwidth]{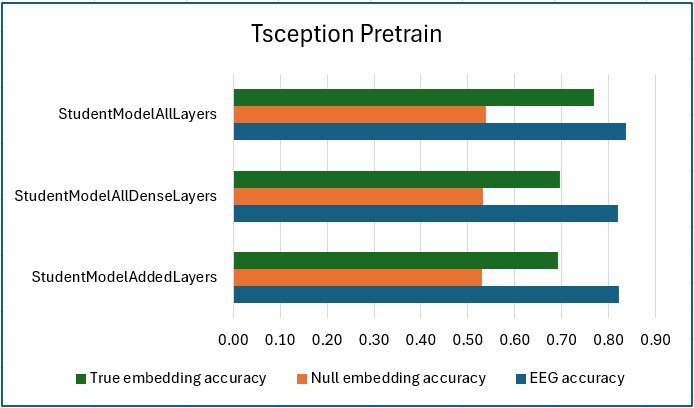}
	\end{subfigure}
	\hfill 
	\caption{The accuracies of the primary task (EEG) and watermark (Null \& True Embedding) when TransferLearning Pretrain models for different four methods.}
	\label{TransferLearningPretrain}
\end{figure}

Transfer learning is shown in tables \ref{TransferLearningFromscratch}, \ref{TransferLearningPretrain} for fromscratch and pretrain models, it is shown that AllLayers methods cause more damage to the watermark, like in fine tuning, compared to the other two settings because they also modify a larger number of model weights during the learning process. There is also ab problem with Tsception fromscratch task as the true and null embedding accuricies dropped to about 60\% which is below the expected value, but the EEG task was also affected by 5\% drop.

\subsubsection{Pruning}
\begin{figure}[htbp]
	\centering
	\begin{subfigure}[b]{0.98\columnwidth}
		\includegraphics[width=\textwidth]{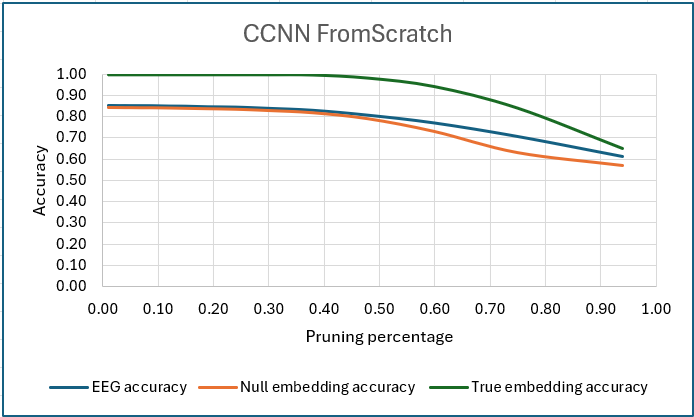}
	\end{subfigure}
	\hfill
	\begin{subfigure}[b]{0.98\columnwidth}
		\includegraphics[width=\textwidth]{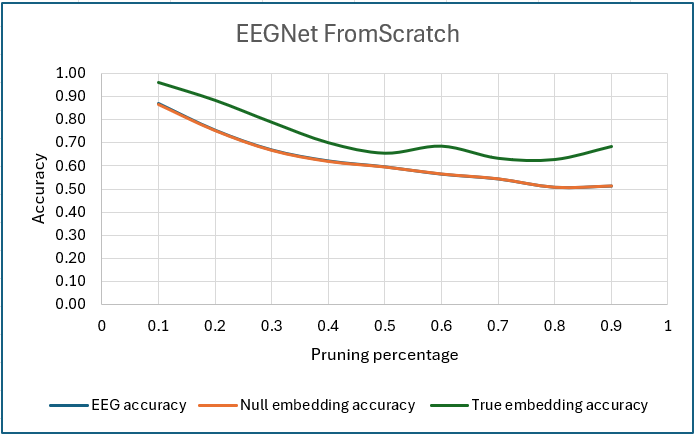}
	\end{subfigure}
	\hfill
	\begin{subfigure}[b]{0.98\columnwidth}
		\includegraphics[width=\textwidth]{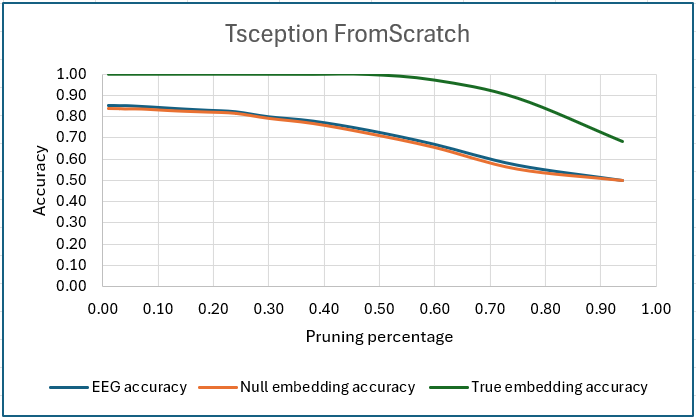}
	\end{subfigure}
	\caption{The accuracies of the primary task (EEG) and watermark (Null \& True Embedding) when pruning FromScratch models with L1Pruning for different ratios of the model's neurons.}
	\label{L1FromScratch}
\end{figure}

\begin{figure}[htbp]
	\centering
	\begin{subfigure}[b]{0.98\columnwidth}
		\includegraphics[width=\textwidth]{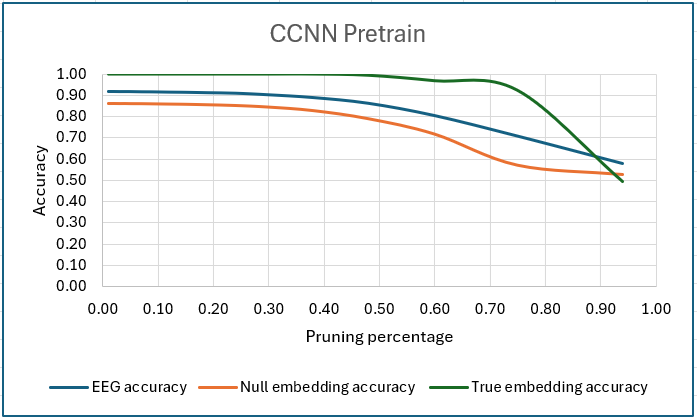}
	\end{subfigure}
	\hfill
	\begin{subfigure}[b]{0.98\columnwidth}
		\includegraphics[width=\textwidth]{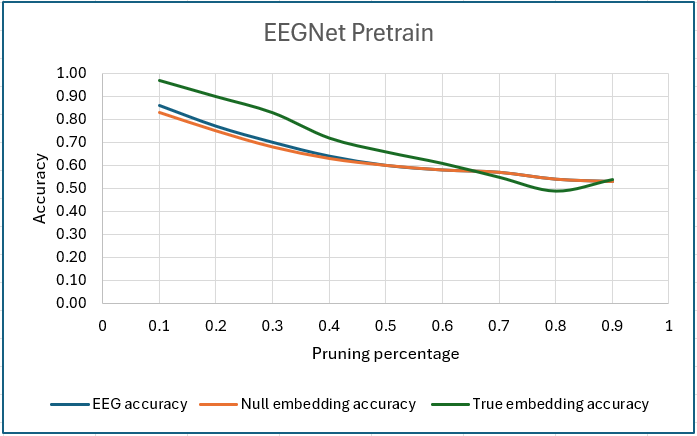}
	\end{subfigure}
	\begin{subfigure}[b]{0.98\columnwidth} 
		\includegraphics[width=\textwidth]{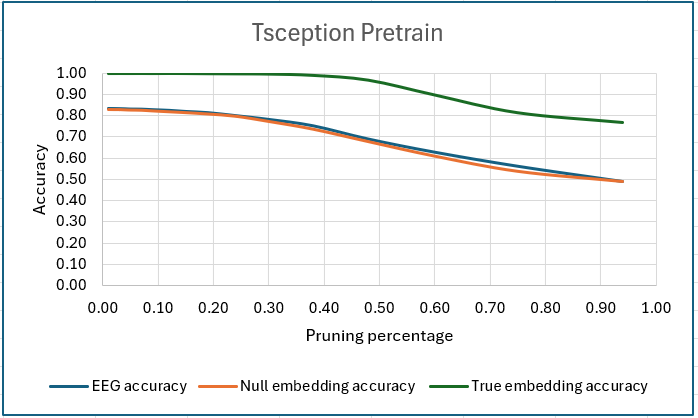}
	\end{subfigure}
	\hfill 
	\caption{The accuracies of the primary task (EEG) and watermark (Null \& True Embedding) when pruning Pretrain models with L1Pruning for different ratios of the model's neurons.}
	\label{L1Pretrain}
\end{figure}

\begin{figure}[htbp]
	\centering
	\begin{subfigure}[b]{0.98\columnwidth}
		\includegraphics[width=\textwidth]{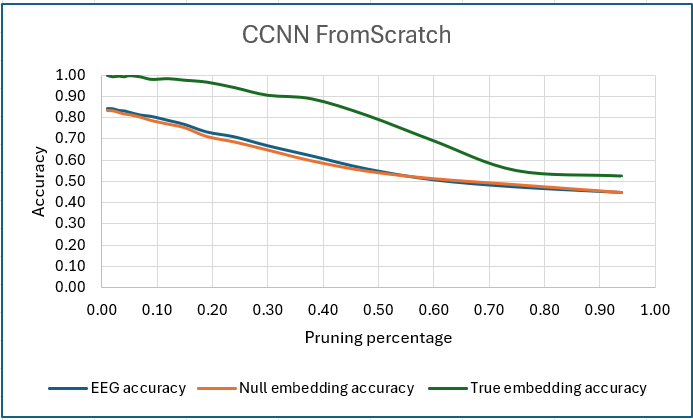}
	\end{subfigure}
	\hfill
	\begin{subfigure}[b]{0.98\columnwidth}
		\includegraphics[width=\textwidth]{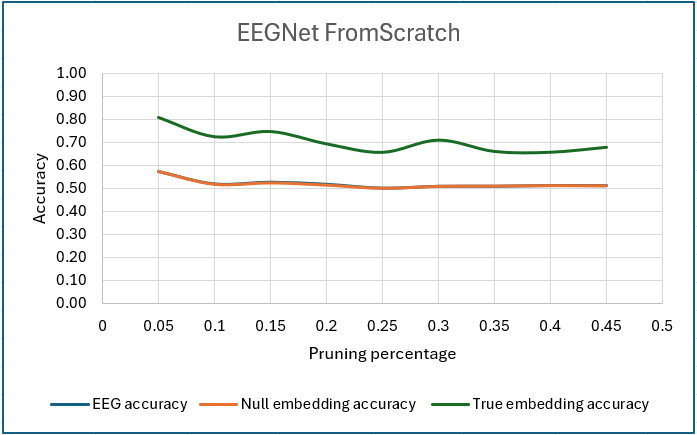}
	\end{subfigure}
	\hfill
	\begin{subfigure}[b]{0.98\columnwidth}
		\includegraphics[width=\textwidth]{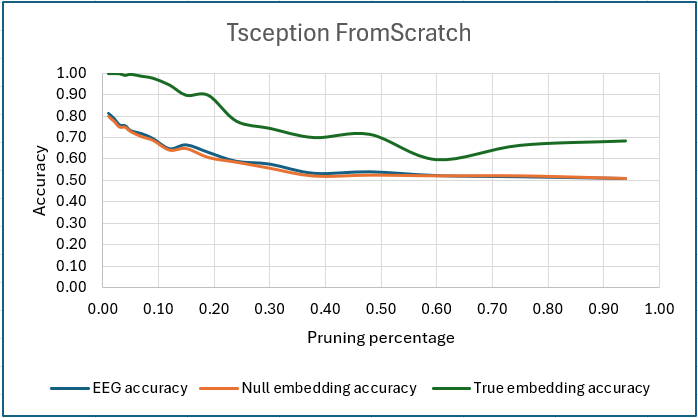}
	\end{subfigure}
	\caption{The accuracies of the primary task (EEG) and watermark (Null \& True Embedding) when pruning FromScratch models with RandomPruning for different ratios of the model's neurons.}
	\label{RandomFromScratch}
\end{figure}

\begin{figure}[htbp]
	\centering
	\begin{subfigure}[b]{0.98\columnwidth}
		\includegraphics[width=\textwidth]{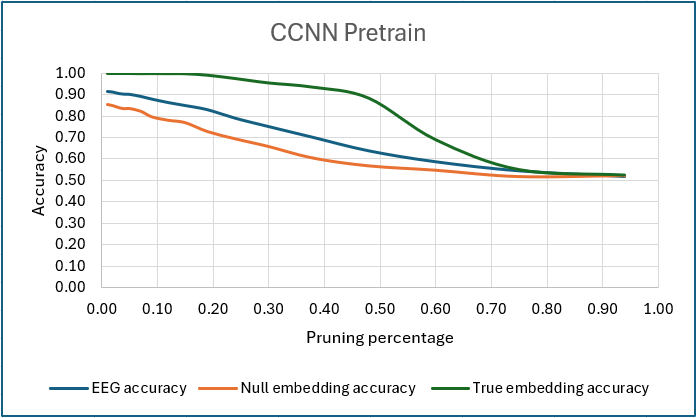}
	\end{subfigure}
	\hfill
	\begin{subfigure}[b]{0.98\columnwidth}
		\includegraphics[width=\textwidth]{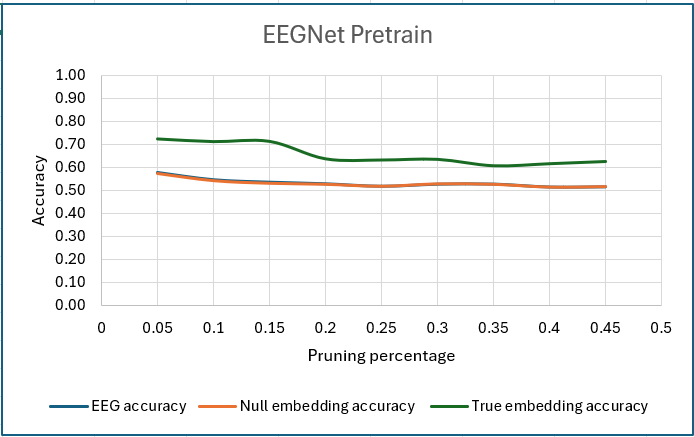}
	\end{subfigure}
	\begin{subfigure}[b]{0.98\columnwidth} 
		\includegraphics[width=\textwidth]{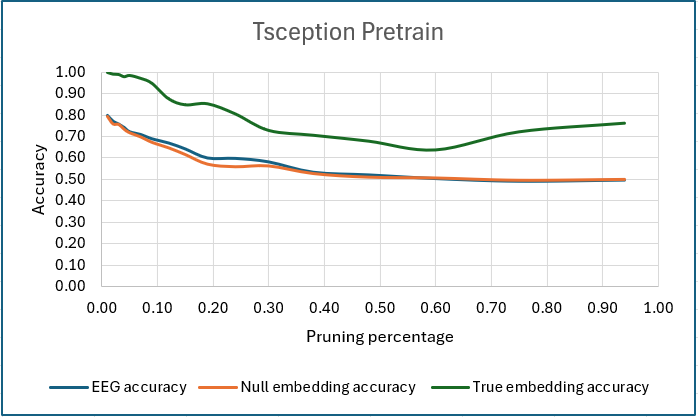}
	\end{subfigure}
	\hfill 
	\caption{The accuracies of the primary task (EEG) and watermark (Null \& True Embedding) when pruning Pretrain models with RandomPruning for different ratios of the model's neurons.}
	\label{RandomPretrain}
\end{figure}

As shown in Figures \ref{L1FromScratch}, \ref{L1Pretrain}, \ref{RandomPretrain}, and \ref{RandomFromScratch} with the ratio of pruned neurons for L1 and Random pruning, as the pruned neurons is increasing, the trend of the accuracy of EEG, null and true embedding is decreasing. That is consistent in the two pruning strategies but with more effect in the random strategy. The above observations show that the neurons activated by watermarked inputs are also activated by EEG inputs. Which once again proves the robustness of the proposed watermarking scheme in persistance.

\subsection{Piracy Resistance}
Table \ref{NewWatermark} shows the accuracy of the attacker's watermark \(W_A\) after embedding it with his limited resources. The EEGNet and CCNN models undergo a significant drop of approximately \(\geq 10\%\), while Tsception was a little less drop in accuracy of approximately 5\%.

\begin{table*}[]
	\centering
	\caption{Classification accuracy of Attacker Watermark \(W_A\) (Null and True embedding sets) and primary task (EEG accuracy) after embedding their malicious watermark, with the two watermarking strategies for each model.}
	\label{NewWatermark}
	\begin{tabular}{ccccc}
		\hline
		\rowcolor[HTML]{FFFFFF} 
		\multicolumn{1}{l}{\cellcolor[HTML]{FFFFFF}{\color[HTML]{000000} \textbf{EEG Model}}} &
		{\color[HTML]{000000} \textbf{Training method}} &
		\multicolumn{1}{l}{\cellcolor[HTML]{FFFFFF}{\color[HTML]{000000} \textbf{EEG accuracy}}} &
		\multicolumn{1}{l}{\cellcolor[HTML]{FFFFFF}{\color[HTML]{000000} \textbf{Null embedding accuracy}}} &
		\multicolumn{1}{l}{\cellcolor[HTML]{FFFFFF}{\color[HTML]{000000} \textbf{True embedding accuracy}}} \\ \hline
		\cellcolor[HTML]{FFFFFF}{\color[HTML]{000000} }                            & From scratch & 0.80 & 0.74 & 1.00 \\ \cline{2-5} 
		\multirow{-2}{*}{\cellcolor[HTML]{FFFFFF}{\color[HTML]{000000} CCNN}}      & Pretrain     & 0.82 & 0.77 & 1.00 \\ \hline
		\cellcolor[HTML]{FFFFFF}{\color[HTML]{000000} }                            & From scratch & 0.74 & 0.75 & 0.99 \\ \cline{2-5} 
		\multirow{-2}{*}{\cellcolor[HTML]{FFFFFF}{\color[HTML]{000000} EEGNet}}    & Pretrain     & 0.76 & 0.75 & 0.99 \\ \hline
		\cellcolor[HTML]{FFFFFF}{\color[HTML]{000000} }                            & From scratch & 0.80 & 0.79 & 0.99 \\ \cline{2-5} 
		\multirow{-2}{*}{\cellcolor[HTML]{FFFFFF}{\color[HTML]{000000} Tsception}} & Pretrain     & 0.80 & 0.79 & 0.99 \\ \hline
	\end{tabular}
\end{table*}

\section{Discussion}
The study begins by verifying that the proposed wonder filter-based watermark meets three fundamental requirements for effective watermarking: low distortion, high reliability, and no false positives. These requirements are critical prerequisites for enabling advanced functionalities such as authentication, persistence, and piracy resistance. If a watermark compromises or affects model accuracy or exhibits inconsistent detection, it can not support these higher-level properties and requirements for successful watermark embedding. 

\subsection{Basic Requirements}

To evaluate these requirements, all models were trained from scratch for each of three distinct tasks (in addition to pretrain strategy and non watermarked model). The results of tables \ref{TriggerSetWatermarking}, \ref{WrongVerifier} has insights about these initial requirements as follows:
\begin{itemize}
	\item Low Distortion: The watermark’s impact on model primary task (EEG accuracy) was very minimal across all tasks and models. As shown in Table \ref{TriggerSetWatermarking}, classification accuracy for watermarked models decreased by about only 5\% compared to their watermark-free counterparts. This negligible drop in the accuracy confirms that the watermark does not interfere with the model’s primary functionality, and ensuring its practicality for real-world deployment and dependency.
	\item High Reliability: The watermarks accuracies in table \ref{TriggerSetWatermarking} demonstrated near-perfect models that are reliabile in detection of the primary task and in the embedded watermark. For all tasks, watermark accuracy was measured under two states: the pretrain embedding and the fromscratch embedding. Normal embedding accuracy reached 100\% in almost all cases, as the wonder pattern’s explicit association with one and only one specific label form the domain of labels, that guarantees easy and unambiguous detection for all models. Null embedding accuracy was almost equal or slightly less than the primary EEG task because their labels are the same. So, the models can learn that the pixels found in the position of the wonder filter has no effect on the classification result.
	      	      	          
	\item No False Positives: The low null and true embedding accuracy (\(>\)50\%) in table \ref{WrongVerifier} reflects an exceptionally nigligible probability of detecting watermarks in non-watermarked models. This specificity is critical for legal and operational credibility, as false positives could lead to incorrect claims of ownership or piracy by attackers. These results are indicating the robust performance in avoiding false identifications when no watermark exists in a model.
\end{itemize}

\subsection{Advanced Requirements}
Now the performance of the watermark scheme is assessed against the advanced criteria outlined in Section II. Prior research has developed watermarks that satisfy these requirements for neural networks that deal with images, but the EEG-based models does not satisfy them, especially authentication and piracy resistance, as attackers often succeed in embedding their own watermarks. The wonder filter is used for EEG-based watermarking as it should meet all advanced requirements and demonstrate strong resilience against piracy attempts.

\subsubsection{Authentication}
This scheme inherently fulfills the authentication requirement and it is extremely critical feature as it can prove the ownership for specific user. By leveraging a collision-resistant hash function (Section IV) to generate the watermark, the likelihood of two distinct watermark masks producing identical hashes is equivalent to an adversary randomly guessing the correct mask. As demonstrated earlier, this probability is extremely low, ensuring the watermark’s role as a secure authentication mechanism.
\subsubsection{Persistence}
To evaluate the watermark’s persistence, three adversarial strategies have been used: model fine-tuning, transfer learning, and neuron pruning. For these tests, it is assumed that the adversary aims to remove the watermark while preserving the model’s primary functionality (i.e., maintaining high classification accuracy).
\begin{itemize}
	\item Fine-Tuning Resistance
	      Fine tuning is a common technique used for updating model weights without degrading accuracy. Tt was applied to all layers of the model. Results in Figure \ref{FineTuningFromScratch}, \ref{FineTuningPretrain} reveal that even after 30 epochs of fine-tuning, the watermark remains intact. While normal classification accuracy slightly dropped due to overfitting on the limited fine-tuning dataset, normal embedding accuracy persists at above 90\%. Null embedding accuracy exhibits more fluctuations but remains stable, confirming the watermark’s robustness. RTAL method updates all weights, so it is stronger than other techniques, but still it has no strong effect except on Tsceptoin pretrain model
	      	      	      
	\item Neuron Pruning Resistance
	      	      	      
	      To test resilience against architectural modifications, an ascending pruning (removing neurons with smaller absolute weights first) test was conducted as shown in figures\ref{L1FromScratch}, \ref{L1Pretrain}. It is a standard method for model compression [9], [10] by canceling weights that might be useless or has little effect on the classification result. The test was conducted on pretrain and fromscratch models. The figures illustrate the effects of varying pruning ratios on classification and watermark accuracy.
	      	      	      
	      For all models, normal classification accuracy degrades faster than watermark accuracy as pruning intensifies. This is a strong indication on the persistance of the watermark, as the avdersary needs to sacrifice the primary task before being able to remove the embedded watermark. This is not reasonable, as the model will be useless if it loses the primary task classification ability.
	      	      	      
	      Another method used is random pruning, fig. \ref{RandomPretrain} and \ref{RandomFromScratch}, which chooses the bruned weights at random, In this method, both normal classification and null embedding accuracy decline sharply in EEGNet, for example, with just 5\% of neurons pruned.
	      	      	      
	      Critically, no pruning level achieves acceptable classification performance while disrupting the watermark.
	      	      	      
\end{itemize}

\subsection{Piracy resistance}
The table \ref{NewWatermark} provides the result for adversary trying to embed new malisious watermark to test whether this watermarking scheme can withstand ownership piracy attacks, where an attacker tries to add their own watermark into the model. 

It was already demonstrated that attackers cannot remove existing watermarks from models using fine tuning or pruning techniques; this is protection against corruption attacks. This also protects the model from the second type of piracy attack, known as takover, because that attack also involves the attacker removing or retraining the owner’s watermark.

Therefore, the focus here is on the third type of attack, simply piracy, where the attacker embeds their own watermark into the model alongside the owner’s watermark. Consider an attacker \(A\) who uses the approach of the watermarking scheme to embed their own watermark, labeled \(W_A\), into the model. The findings in table \ref{WrongVerifier} indicate that it is difficult to embed a new watermark on top of an existing one. The table compares the normal classification accuracy and watermark accuracies of a model after the attacker attempts to embed a new watermark. After the embedding process, the accuracy of primary task and null embedding was affected with average of about 10\% in EEGNet and CCNN models which is high and critical drop. For the Tsception, it was about 5\% which is not enough to indicate success. But, the original watermark \(W\) is still exixsing in the model and the owner can provide a model with \(W\) only without \(W_A\) embedded and that can be a strong evidence for his ownership \cite{Zhu2020}. The attacker will have a model with the two watermarks \(W\) and \(W_A\) embedded, so this can provide reasonable claim for legitimate owner, as it is impossible that the owner removed \(W_A\) from his model. These results and analysis show that it is very difficult to embed a new watermark into a model that has already been watermarked.

Even after attempting to embed the second watermark for more than 100 epochs, the attacker makes no progress. Training for the new watermark WA completely fails, and both the normal classification accuracy and the owner’s watermark remain completely unaffected. This confirms our claim that watermark training can only be successfully completed during model training time, and that our watermark system effectively resists ownership piracy attacks.

\section{Conclusion}

This study addressed the critical challenge of protecting EEG-based neural networks, which are vital for medical diagnosis and brain-computer interfaces. Traditional watermarking methods, reliant on abstract triggers, often lack robust authentication and fail to meet the unique demands of EEG models. The proposed wonder filter-based framework, integrated with cryptographic techniques, successfully embeds a secure, tamper-proof watermark while preserving model functionality. Key findings demonstrate minimal performance degradation, resilience against adversarial attacks, and unparalleled authentication, fulfilling the objectives of persistence, piracy resistance, and ownership verification.

Key Findings:
The wonder filter-based watermarking method achieved three foundational goals:
\begin{itemize}
    \item Minimal Distortion: Embedding the watermark caused less than a 5\% drop in EEG classification accuracy across models (CCNN, EEGNet, TSception), ensuring practical utility in real-world applications like emotion recognition and neurological disorder detection.

    \item Robust Persistence: The watermark resisted removal through aggressive fine-tuning (100 epochs) and neuron pruning (up to 50\%). Notably, the primary task accuracy degraded faster than watermark detection, rendering adversarial attempts to remove the mark counterproductive.

    \item Piracy Resistance: Attempts to embed secondary watermarks caused significant accuracy losses (\(>10\%\) in EEGNet and CCNN), deterring unauthorized claims of ownership. Cryptographic hashing, tied to the owner’s private key, reduced brute-force attack success probabilities, making such attacks computationally infeasible.
\end{itemize}

The study aimed to adapt wonder filters for EEG models and address three advanced requirements: authentication, persistence, and piracy resistance. Cryptographic hashing ensured authentication by linking the watermark to the owner’s unique signature, eliminating reliance on easily replicated abstract triggers. Persistence was validated through rigorous testing against fine-tuning and pruning, common post-deployment modifications. Piracy resistance was demonstrated by the inability to embed competing watermarks without degrading model functionality, fulfilling the objective of securing exclusive ownership.

This work makes four key contributions to the field of neural network watermarking:
\begin{itemize}
	\item EEG-Specific Adaptation: The wonder filter, previously applied to image-based models, was successfully tailored to EEG architectures, addressing the unique challenges of neurophysiological data, such as low signal-to-noise ratios and temporal complexity.
	      	      
	\item Cryptographic Authentication: By integrating digital signatures and collision-resistant hashing, the framework provides irrefutable proof of ownership, overcoming the authentication weaknesses of prior EEG watermarking methods.
	      	      
	\item Attack Resilience: The watermark’s robustness against fine-tuning, pruning, and adversarial overwriting sets a new benchmark for securing EEG models in high-stakes environments.
	      	          
	\item Practical Security: The method balances security with usability, ensuring minimal impact on model performance—a critical requirement for medical and biometric applications where accuracy is paramount.
	      	          
\end{itemize}

This framework offers a reliable solution for safeguarding EEG-based AI systems, which are increasingly deployed in healthcare, neuroscience, and biometrics. By preventing unauthorized use and tampering, it encourages innovation and collaboration while protecting investments in resource-intensive model development. For instance, hospitals using watermarked EEG models for seizure detection can confidently share tools without risking intellectual property theft. Similarly, developers of brain-computer interfaces can license their models securely, knowing ownership claims are cryptographically verifiable.

While the method excels in current tests, extending it to larger, more complex EEG architectures and diverse applications (e.g., Alzheimer’s detection, sleep monitoring) will further validate its versatility. Investigating hybrid approaches—combining wonder filters with encryption or secure hardware—could enhance protection against emerging attack vectors. Additionally, real-world deployment studies are needed to assess long-term robustness in dynamic clinical environments.

In summary, this study advances the protection of EEG-based neural networks by introducing a watermarking framework that combines cryptographic security with practical resilience. By addressing the limitations of prior methods and aligning with the unique demands of neurophysiological data, the work supports the ethical and secure advancement of AI in sensitive, life-critical domains.



%






%

\bibliographystyle{IEEEtran}
\bibliography{references}

\onecolumn
\newpage
\appendix

\begin{table}[htbp]
	\centering
    \caption*{CCNN Architecture}
    	\begin{adjustbox}{width=1\textwidth}
		\def\arraystretch{1.25}
		\begin{tabular}{c|lll}
			\hline
			Model Structure & Layers                                      & Input             & Output            \\
			\hline
			Block1          & ZeroPad2d(\com{(1, 2, 1, 2)})               & (-1, 4, 9, 9)     & (-1, 4, 12, 12)   \\
			                & Conv2d(\com{4, 64, kernel\_size=(4, 4)})    & (-1, 4, 12, 12)   & (-1, 64, 9, 9)    \\
			                & ReLU()                                      & (-1, 64, 9, 9)    & (-1, 64, 9, 9)    \\ \hline
			Block2          & ZeroPad2d(\com{(1, 2, 1, 2)})               & (-1, 64, 9, 9)    & (-1, 64, 12, 12)  \\
			                & Conv2d(\com{64, 128, kernel\_size=(4, 4)})  & (-1, 64, 12, 12)  & (-1, 128, 9, 9)   \\
			                & ReLU()                                      & (-1, 128, 9, 9)   & (-1, 128, 9, 9)   \\ \hline
			Block3          & ZeroPad2d(\com{(1, 2, 1, 2)})               & (-1, 128, 9, 9)   & (-1, 128, 12, 12) \\
			                & Conv2d(\com{128, 256, kernel\_size=(4, 4)}) & (-1, 128, 12, 12) & (-1, 256, 9, 9)   \\
			                & ReLU()                                      & (-1, 256, 9, 9)   & (-1, 256, 9, 9)   \\ \hline
			Block4          & ZeroPad2d(\com{(1, 2, 1, 2)})               & (-1, 256, 9, 9)   & (-1, 256, 12, 12) \\
			                & Conv2d(\com{256, 64, kernel\_size=(4, 4)})  & (-1, 256, 12, 12) & (-1, 64, 9, 9)    \\
			                & ReLU()                                      & (-1, 64, 9, 9)    & (-1, 64, 9, 9)    \\ \hline
			Fully           & Flatten                                     & (-1, 64, 9, 9)    & (-1, 5184)        \\
			Connected       & Linear(\com{5184, 1024})                    & (-1, 5184)        & (-1, 1024)        \\
			Layers          & SELU()                                      & (-1, 1024)        & (-1, 1024)        \\
			                & Dropout2d(\com{p=0.5})                      & (-1, 1024)        & (-1, 1024)        \\
			                & Linear(\com{1024, 2})                       & (-1, 1024)        & (-1, 2)           \\
			                & Softmax                                     & (-1, 2)           & (-1, 2)           \\
			\hline
		\end{tabular}
	\end{adjustbox}
	\vspace{3mm}
	\caption{CCNN architecture: ZeroPad2d adds padding to the input tensor. Conv2d applies 2D convolution, ReLU and SELU are activation functions, and Dropout2d applies dropout regularization. The input tensor shape is (-1, 4, 9, 9), where -1 represents the batch size, 4 is the number of input channels, and (9, 9) is the spatial dimension. The output tensor shape is (-1, 2), where 2 corresponds to the number of classes.\cite{yang2018ccnn}} \label{Tab:ccnn_architecture}
\end{table}

\begin{table}[htbp]
	\centering
    \caption*{TSCeption Architecture}
	\begin{adjustbox}{width=1\textwidth}
						 	
		\def\arraystretch{1.25}
		\begin{tabular}{cc|llllllll}
			\hline
			Model structure &               & Layers                                        & Input             & Output            \\
			\hline
			Block1          & 3 branches    & Conv2d, LK-ReLU, AP(\com{(1,8)})              & (-1, 1, 28, 512)  & (-1, 15, 28, 56)  \\
			                & (in parallel) & Kernel=\com{15@(1, 64)}                       &                   &                   \\ \cline{3-5}
			                &               & Conv2d, LK-ReLU, AP(\com{(1,8)})              & (-1, 1, 28, 512)  & (-1, 15, 28, 60)  \\
			                &               & Kernel=\com{15@(1, 32)}                       &                   &                   \\ \cline{3-5}
			                &               & Conv2d, LK-ReLU, AP(\com{(1,8)})              & (-1, 1, 28, 512)  & (-1, 15, 28, 62)  \\
			                &               & Kernel=\com{15@(1, 16)}                       &                   &                   \\ \cline{2-5}
			                &               & Concatenate, BN                               &                   & (-1, 15, 28, 178) \\
			\hline
			Block2          & 2 branches    & Conv2d, LK-ReLU, AP(\com{(1,2)})              & (-1, 15, 28, 178) & (-1, 15, 1, 89)   \\
			                & (in parallel) & Kernel=\com{15@(28, 1)}                       &                   &                   \\ \cline{3-5}
			                &               & Conv2d, LK-ReLU, AP(\com{(1,2)})              & (-1, 15, 28, 178) & (-1, 15, 2, 89)   \\
			                &               & Kernel=\com{15@(14, 1)}, Stride=\com{(14, 1)} &                   &                   \\ \cline{2-5}
			                &               & Concatenate, BN                               &                   & (-1, 15, 3, 89)   \\
			\hline
			Block3          &               & Conv2d, LK-ReLU, AP(\com{(1,4)}), BN, GAP     & (-1, 15, 3, 89)   & (-1, 15, 1)       \\
			                &               & Kernel=\com{15@(3, 1)}                        &                   &                   \\ \cline{3-5}
			                &               & Flatten                                       & (-1, 15, 1)       & (-1, 15,)         \\
			\hline
			Fully           &               & Linear(\com{32}), ReLU                        & (-1, 15,)         & (-1, 32,)         \\
			Connected       &               & dropout(\com{0.5})                            & (-1, 32,)         & (-1, 32,)         \\
			Layers          &               & Linear(\com{2})                               & (-1, 32,)         & (-1, 2,)          \\
			                &               & softmax                                       & (-1, 2,)          & (-1, 2,)          \\
			\hline
		\end{tabular}
	\end{adjustbox}
	\vspace{3mm}
	\caption{TSCeption architecture: LK-ReLU is the Leaky-ReLU activation function. AP is the average pooling operation. BN stands for batch normalization. GAP is the global average pooling. '-1' in the tensor size stands for the number of samples within one mini-batch. The strides of CNNs are (1, 1) if not specified, and the one for pooling layers is the same as the pooling step.\cite{ding2023tsception}} \label{Tab:tsception_architecture}
\end{table}

\begin{table}[htbp]
	\centering
    \caption*{EEGNet Architecture}
	\begin{adjustbox}{width=1\textwidth}
						 	
		\def\arraystretch{1.25}
		\begin{tabular}{c|lllllllll}
			\textbf{Block} & \textbf{Layer}  & \textbf{\# filters}   & \textbf{size} & \textbf{\# params}               & \textbf{Output}        & \textbf{Activation} \\ \hline
									 		  
			1              & Input           &                       &               &                                  & (C, T)                 &                     \\
			               & Reshape         &                       &               &                                  & (1, C, T)              &                     \\
			               & Conv2D          & $F_1$                 & (1, 64)       & $64 * F_1$                       & ($F_1$, C, T)          & Linear              \\
			               & BatchNorm       &                       &               & $2 * F_1$                        & ($F_1$, C, T)          &                     \\
			               & DepthwiseConv2D & D * $F_1$             & (C, 1)        & $C * D * F_1$                    & (D * $F_1$, 1, T)      & Linear              \\
			               & BatchNorm       &                       &               & $2 * D * F_1$                    & (D * $F_1$, 1, T)      &                     \\
			               & Activation      &                       &               &                                  & (D * $F_1$, 1, T)      & ELU                 \\
			               & AveragePool2D   &                       & (1, 4)        &                                  & (D * $F_1$, 1, T // 4) &                     \\
			               & Dropout*        &                       &               &                                  & (D * $F_1$, 1, T // 4) &                     \\
			2              & SeparableConv2D & $F_2$                 & (1, 16)       & $16 * D * F_1 + F_2 * (D * F_1)$ & ($F_2$, 1, T // 4)     & Linear              \\
			               & BatchNorm       &                       &               & $2 * F_2$                        & ($F_2$, 1, T // 4)     &                     \\
			               & Activation      &                       &               &                                  & ($F_2$, 1, T // 4)     & ELU                 \\
			               & AveragePool2D   &                       & (1, 8)        &                                  & ($F_2$, 1, T // 32)    &                     \\
			               & Dropout*        &                       &               &                                  & ($F_2$, 1, T // 32)    &                     \\
			               & Flatten         &                       &               &                                  & ($F_2$ * (T // 32))    &                     \\
			Classifier     & Dense           & N * ($F_2$ * T // 32) &               &                                  & N                      & Softmax             
		\end{tabular}
	\end{adjustbox}
	\vspace{3mm}
	\caption{EEGNet architecture, where $C = $ number of channels,  $T = $ number of time points, $F_1 = $ number of temporal filters, $D = $ depth multiplier (number of spatial filters), $F_2 = $ number of pointwise filters, and $N = $ number of classes, respectively.\cite{lawhern2018eegnet}} \label{Tab:eegnet_architecture}
\end{table}

\end{document}